# Exploring Advanced Techniques for Visual Question Answering: A Comprehensive Comparison

Aiswarya Baby, Tintu Thankom Koshy

Yeates School of Graduate and Postdoctoral Studies, Toronto Metropolitan University, Canada

## ABSTRACT

Visual Question Answering (VQA) has emerged as a pivotal task in the intersection of computer vision and natural language processing, requiring models to understand and reason about visual content in response to natural language questions. Analyzing VQA datasets is essential for developing robust models that can handle the complexities of multimodal reasoning. Several approaches have been developed to examine these datasets, each offering distinct perspectives on question diversity, answer distribution, and visual-textual correlations. Despite significant progress, existing VQA models face challenges related to dataset bias, limited model complexity, commonsense reasoning gaps, rigid evaluation methods, and generalization to realworld scenarios. This paper offers a detailed study of the original VQA dataset, baseline models and methods along with a comparative study of five advanced VQA models, ABC-CNN, KICNLE, Masked Vision and Language Modeling, BLIP-2, and OFA, each employing distinct methods to address these ongoing challenges.

## 1. INTRODUCTION

Visual Question Answering (VQA) is a complex AI task that merges computer vision and natural language processing (NLP) to enable machines to answer questions based on the content of images. Unlike traditional computer vision tasks such as image classification or object detection, VQA requires models to simultaneously process both visual and textual information, making it a central challenge at the intersection of vision and language [1]. A VQA system must not only identify objects, scenes, and attributes in an image but also interpret the semantics of the question, apply reasoning, and generate a relevant response.

The task has significantly advanced with the rise of deep learning. Early approaches utilized convolutional neural networks (CNNs) for image feature extraction and recurrent neural networks (RNNs) for encoding questions. However, these early models often relied on dataset biases rather than truly understanding image content. For instance, in the VQA dataset, the answer "tennis" was often the correct response to the question "What sport is being played?" without requiring real image analysis. To mitigate such biases, the VQA v2.0 dataset was introduced, providing balanced question-image pairs where answers vary depending on the image. Furthermore, attention mechanisms have been integrated into VQA models to enhance their focus on relevant regions of the image when generating answers.

Analyzing VQA datasets is crucial for developing robust models capable of handling the complexities of multimodal reasoning. Various approaches have been developed to study these datasets, offering different insights into question diversity, answer distribution, and visual-textual correlations. Despite significant progress, existing VQA models still face challenges such as dataset bias, limited model complexity, gaps in commonsense reasoning, rigid evaluation methods, and difficulties in generalizing to real-world scenarios. This paper offers a detailed study of the original VQA dataset, baseline models and methods along with a comparative study of five advanced VQA models, ABC-CNN, KICNLE, Masked Vision and Language Modeling, BLIP-2, and OFA, each employing distinct methods to address these ongoing challenges.

## 2. REVIEW OF EXISTING WORK

Visual Question Answering (VQA) is an example of a multi-modal Visual-Language (VL) task that combines computer vision and natural language processing techniques to enable machines to answer questions about visual content. For instance, given an image of "The Simpsons" TV show and a natural language question such as "Does Homer wear a hat?", a VQA system should be able to recognize "Homer" in the image and generate the correct answer, "No". Supported by various large-scale datasets VQA is widely recognized as one of the most challenging tasks in the field of VL computing, particularly when compared to other VL tasks like Visual Grounding and Visual Captioning. One of the reasons for this is that VQA's input question form is much more diverse than other VL tasks. While the text queries in other tasks like Visual Grounding are fixed and describe specific objects or features in an image, VQA questions can vary widely in format and content. Several surveys on VQA have been published in recent years to review and evaluate different aspects of VQA. However, with the rapid progress of the field, combined with the introduction of new techniques, models, and datasets, there is a need for an up-to-date and comprehensive survey on the topic.

### 2.1. Foundational Work: The VQA Dataset, Baseline Model and Methods

The paper "VQA: Visual Question Answering" [1] introduced a multimodal task combining computer vision and natural language processing to answer open-ended questions about images. It is deemed "AI-complete" as it requires visual, linguistic, and commonsense reasoning. Addressing prior limitations of template-based questions and small datasets, VQA uses free-form, human-like questions (e.g., "How many chairs are in the room?"). A large-scale dataset was curated with 204,721 real images (MS COCO), 50,000 synthetic scenes, more than 760,000 questions, and 10 million answers. The questions were designed to challenge a smart robot capable of understanding images and answering basic questions. They were crowdsourced via Amazon Mechanical Turk to ensure diversity across different question types (yes/no, counting, and reasoning). Answers were aggregated from 10 subjects along with confidence scores, validated by consensus (≥3/10 agreement). MS COCO images included five captions each, while abstract image captions were manually created by the authors. Evaluation was conducted using open-response (exact match) and multiple-choice (18-option) formats, balancing real-world applicability with automated scoring.

The authors introduced several baseline models and deep learning approaches to evaluate performance. The baseline models included a random answer selection, a prior-based model that always predicted "yes" (the most common answer), and a per-question-type prior, which selected the most frequent answer for each specific question type. A nearest neighbor approach was also implemented, where the model identified similar training questions and selected the most common ground truth answer. These simple benchmarks highlighted the difficulty of the VQA task and the need for more sophisticated methods that integrate both vision and language understanding.

They proposed a two-channel vision-language deep learning model that predicts answers using a softmax over the 1,000 most frequent answers, covering 82.67% of the train + val dataset. The image channel extracts 4096-dimensional VGGNet features, either directly (I) or after L2 normalization (norm I). The question channel utilizes three embeddings: Bag-of-Words (BoW Q), a one-layer LSTM (1024-dim), and a two-layer LSTM (2048-dim). Embeddings are fused through concatenation (BoW Q + I) or element-wise multiplication (LSTM Q + I, deeper LSTM Q + norm I) before being passed through a Multi-Layer Perceptron (MLP) with two hidden layers (1000 units each, dropout 0.5, tanh activation). The final softmax layer generates predictions over the top 1,000 answers, ensuring a balance between coverage and efficiency. The model is trained using cross-entropy loss, with VGGNet weights frozen to prevent overfitting. The best-performing model, deeper LSTM Q + norm I, achieved 58.16% accuracy (open-ended) and 63.09% (multiple-choice). While significantly outperforming baselines, it remains below human accuracy (80.62%), highlighting the need for better reasoning and external knowledge integration.

Strengths of this model include its ability to handle diverse question types and its robust fusion mechanism, which effectively combines visual and textual cues. The use of normalized image features and deeper LSTMs enhances stability and reasoning depth. However, weaknesses persist: the model relies heavily on frequent answers (e.g., "yes," "2"), struggles with rare or ambiguous answers, and fails to capture nuanced reasoning (e.g., "Why is the woman holding an umbrella?"). The evaluation metric (for example, exact string matching) also penalizes valid synonyms (e.g., "happy" vs. "joyful"). Compared to human performance (83.3% for real images), the model's gaps underscore the complexity of true visual understanding. Despite these limitations, the architecture sets a strong baseline for VQA, emphasizing the need for future work on commonsense integration, robustness to rare answers, and more flexible evaluation frameworks.

## 2.2. Methodological Advancements in VQA

### 2.2.1. ABC-CNN: Attention-Based Configurable Convolutional Neural Network

The Attention-Based Configurable Convolutional Neural Network (ABC-CNN) [2] is designed to improve Visual Question Answering (VQA) by focusing on the most relevant regions of an image based on the input question. Unlike traditional VQA models that process entire images uniformly, potentially overlooking crucial details, ABC-CNN uses question-guided attention to focus on relevant image regions before answering. This attention mechanism enhances both interpretability and answer accuracy. ABC-CNN's architecture consist of four key components:

1. Image Feature Extraction Module: Uses VGG-19 deep CNN to generate spatial feature maps instead of global features, preserving spatial information.
2. Question Understanding Module: Employs an LSTM network to generate dense question embeddings.
3. Attention Extraction Module: Converts question embeddings into configurable convolution kernels (CCK), which are then convolved with the image feature map to generate a Question-Guided Attention Map (QAM).
4. Answer Generation Module: Uses a multi-class classifier trained on attention-weighted image features and question embeddings to predict a single-word answer.

ABC-CNN is evaluated on Toronto COCO-QA, DAQUAR, and VQA datasets, achieving state-of-the-art performance. It attained 58.03% accuracy on Toronto COCO-QA, 42.76% on DAQUAR-reduced, 25.37% on DAQUAR-full, and 48.38% on VQA (evaluated on the top 1,000 most frequent answers). A key strength of ABC-CNN is its question-guided attention mechanism, which allows the model to focus on relevant image regions, leading to improved accuracy and interpretability. The fully convolutional model (ATT-SEG) enhances efficiency by extracting feature maps faster than ATT, while combining both models (ATT-VGG-SEG) achieves the best performance, particularly on location-based questions. Ablation studies confirm the critical role of attention, showing a 1.34% accuracy drop without it, and visualizations of Question-Guided Attention Map (QAM)s demonstrate that the model focuses on relevant image regions, improving interpretability. Another advantage is that ABC-CNN does not require manual attention annotations, reducing the need for additional training data.

Despite its strengths, ABC-CNN has several limitations. It is restricted to single-word answers, limiting its real-world applicability for complex responses. The dynamic attention mechanism introduces computational overhead, affecting scalability. The model is also prone to overfitting, particularly with small or biased datasets (e.g., DAQUAR-reduced), and its reliance on fixed pretrained visual features restricts task-specific adaptability. Additionally, it struggles with ambiguous queries involving multiple regions, and its evaluations lack robustness against adversarial or novel inputs. While attention maps improve interpretability, their alignment with human reasoning is unverified, raising concerns about true transparency.

### 2.2.2. Knowledge-Augmented Visual Question Answering with Natural Language Explanation (KICNLE)

The KICNLE [3] model addresses two major VQA challenges: ensuring consistency between answers and explanations and aligning visual and textual modalities. Traditional models often neglect explanations or lack external knowledge, leading to incoherent responses. KICNLE combines iterative refinement with knowledge augmentation, where explanations refine answers and vice versa, ensuring mutual improvement. The model is trained using Distilled GPT-2 to generate answers and explanations. Adam optimizer is used for training, and hyperparameters are fine-tuned to balance losses during answer and explanation generation. The KICNLE model consists of three key components:

1. Original Information Extractor: Uses Vision Transformer (ViT) to extract patch-level image features and Oscar to obtain multimodal representations aligning images and questions.
2. Knowledge Retrieval Module (KRM): Uses ConceptNet to retrieve external knowledge relevant to the image and question, bridging the semantic gap.
3. Iterative Consensus Generator: Alternates between generating rough answers and corresponding explanations, refining both through a multi-iteration feedback method.

The model is evaluated on three datasets:

- QA-X: A subset of VQA v2, containing question-answer pairs with human-written explanations
- A-OKVQA: Focuses on commonsense reasoning and external knowledge.
- VQA-X-KB: Extension of VQA-X dataset, integrating additional knowledge from ConceptNet to bridge the gap between question and image

KICNLE achieves state-of-the-art performance in VQA-NLE, improving both answer accuracy and explanation quality. On VQA-X, it outperforms previous models with a +3.3% BLEU-4 and +1.8% CIDEr gain, and a 3.65-point accuracy increase over NLX-GPT. On A-OKVQA, it surpasses KRISP by 9.97 points and NLX-GPT by 3.92 points. Ablation studies highlight the importance of iterative refinement and knowledge retrieval, as removing the Knowledge Retrieval Module (KRM) drops CIDEr scores by 10.46 points and reduces explanation quality.

KICNLE's iterative consensus mechanism improves answer-explanation consistency, making responses more coherent and accurate. External knowledge integration from ConceptNet enhances commonsense reasoning, while CLIP and Oscar encoders improve multimodal alignment. Human evaluations confirm that KICNLE generates more fluent and contextually relevant explanations than competing models. It also outperforms NLX-GPT and KRISP, setting a new benchmark for knowledge-augmented VQA.

There are some limitations for the model. The iterative process increases computational cost, making KICNLE slower than single-pass models. Its reliance on ConceptNet may introduce biases and lack domain-specific knowledge. The model also struggles with ambiguous questions and adversarial inputs, limiting robustness. While KICNLE explains answers effectively, its explanations lack deep causal reasoning, affecting trustworthiness. Future improvements should focus on reducing computational overhead, expanding knowledge sources, and improving causal reasoning.

### 2.2.3. Masked Vision and Language Modeling for Multi-Modal Representation Learning

The authors propose a joint masked vision and language (V+L) modeling [4] approach where one modality, either vision (images) or language (text), is reconstructed using masked inputs and corresponding unmasked signals from the other modality. The vision modality consists of image patches extracted using a transformer-based encoder, where certain regions are masked and predicted using text information. The language modality represents text tokens, where specific words or phrases are masked and reconstructed using visual context. The model architecture incorporates cross-attention layers to ensure that both modalities influence the final representation, allowing the system to leverage both vision and language for enhanced understanding.

By modeling $p(T|I)$ (text given image) and $p(I|T)$ (image given text), the model achieves balanced bidirectional alignment, improving performance on tasks like VQA, image-text retrieval, and multimodal generation. The MaskVLM model is trained and evaluated on large-scale vision-language datasets, including MS COCO Captions, Conceptual Captions (CC), and Visual Genome (VG). Notably, MaskVLM outperforms prior models while requiring 40% less training data, demonstrating its data efficiency.

For image-text retrieval, MaskVLM achieves 60.1% R@1 (image) and 83.6% R@1 (text) on MS COCO, and 75.0% R@1 (image) and 92.5% R@1 (text) on Flickr30k, surpassing similar models. In VQA v2, it attains 75.4% accuracy, proving strong vision-language reasoning. It also scores 81.5% on NLVR2 and 80.7% on SNLI-VE, confirming its ability to handle complex image-text relationships. Ablation studies show that removing Masked Image Modeling (MIM) or Masked Language Modeling (MLM) significantly lowers performance, while eliminating cross-attention layers reduces retrieval accuracy by 6.8%, highlighting their importance.

MaskVLM has several key advantages, including end-to-end training, where vision and language representations are jointly learned, data efficiency, allowing it to achieve state-of-the-art results with 40% less training data, and a probabilistic framework that enhances interpretability and robustness. However, it has high computational complexity due to cross-attention mechanisms, making it less scalable than contrastive-only models. It also relies on pre-trained models (ImageNet ViT, RoBERTa), raising concerns about generalization. Additionally, masking strategies and cross-attention depth lack detailed ablation studies, and the distinct contributions of MIM vs. MLM remain unclear despite strong results with limited data.

### 2.2.4. BLIP-2: Bootstrapping Language-Image Pre-training with Frozen Image Encoders and Large Language Models

BLIP-2 [5] introduces a compute-efficient vision-language pretraining strategy by leveraging frozen pre-trained image encoders and frozen large language models (LLMs). This approach significantly reduces computational costs while achieving state-of-the-art performance across various downstream tasks like VQA, image captioning, and image-text retrieval. It introduces a Querying Transformer (Q-Former), which bridges the modality gap between vision and language. Pre-training occurs in two stages: (1) vision-language representation learning, where Q-Former extracts visual features relevant to text, and (2) vision-to-language generative learning, where the extracted features are aligned with a frozen LLM for text generation.

The model is pre-trained using 129 million image-text pairs from datasets including COCO, Visual Genome, CC3M, CC12M, SBU, and LAION400M. Synthetic captions were generated for web images using BLIP-based models. For fine-tuning, datasets such as VQAv2, GQA, OK-VQA, COCO Captioning, and NoCaps were used.

BLIP-2 is computationally efficient. Its modular design enables zero-shot learning and outperforms larger models like Flamingo 80B with 8.7% higher accuracy on VQAv2 while using 54x fewer trainable parameters. Additionally, it excels in image captioning (e.g., COCO, NoCaps) and image-text retrieval (e.g., Flickr30K). Its plug-and-play architecture allows easy integration of newer vision or language models, enhancing future capabilities. It excels in zero-shot instruction-following, handling unseen tasks without fine-tuning, and seamlessly integrates vision-language understanding and generation, proving that high performance can be achieved with minimal computation.

However, BLIP-2 struggles with in-context learning, as it was trained on single image-text pairs, limiting few-shot adaptability. It occasionally generates outdated or inaccurate outputs (e.g., misattributed quotes, obsolete product details) and has challenges with contextual reasoning. Additionally, while efficient, its smaller LLMs underperform on knowledge-intensive tasks like OK-VQA compared to larger models such as Flamingo 80B, highlighting a trade-off between efficiency and knowledge depth. Moreover, inherited biases and risks from frozen LLMs require careful mitigation.

### 2.2.5. OFA: Unifying Architectures, Tasks, and Modalities through a Simple Sequence-to-Sequence Framework

OFA (One For All)[6] proposes a unified sequence-to-sequence (Seq2Seq) architecture that unifies vision, language, and cross-modal tasks (e.g., VQA, image captioning, text-to-image generation) into a single sequence-to-sequence model, eliminating the need for task-specific modules. Built on a Transformer encoder-decoder framework, OFA processes all inputs (text, images, bounding boxes) through a shared vocabulary, enabling multitask learning without task-specific heads or architectural modifications. Key innovations include:

- Unified Representation: Images are discretized into sparse codes via VQGAN, while text uses byte-pair encoding (BPE).
- Instruction-Based Learning: Tasks are formulated as a unified sequence-to-sequence abstraction using handcrafted instructions, allowing the model to perform diverse tasks across modalities without task-specific components.
- Trie-Based Inference: The answers generated by OFA are constrained to a predefined candidate set, improving efficiency and accuracy in classification tasks.

Trained on just 20M image-text pairs, far fewer than SimVLM (1.8B pairs), OFA achieves state-of-the-art performance, including 82.0% accuracy on VQA, 154.9 CIDEr in image captioning, and FID 10.5 in image generation. It also matches specialized unimodal models, reaching 85.6% accuracy on ImageNet-1K. The instruction-based paradigm enables zero-shot generalization, allowing OFA to adapt to unseen tasks and domains (e.g., anime imagery), while Trie-based decoding enhances output reliability. OFA simplifies multimodal AI by combining high performance, data efficiency, and flexibility into a unified framework. Despite its strengths, OFA has some drawbacks. Its zero-shot performance is highly sensitive to instruction phrasing, requiring careful prompt engineering. It struggles with sentence-pair classification tasks (e.g., <60% accuracy on MNLI) due to insufficient pretraining on paired textual data. Multitask interference can occasionally degrade performance, for instance, image infilling improves text-to-image generation but negatively impacts image captioning. Although OFA's data efficiency is impressive, scaling beyond 20M pairs could improve robustness. The model also faces challenges in extreme out-of-domain scenarios, though initial results indicate promising task transfer. Overall, OFA advances multimodal unification, balancing simplicity, efficiency, and high performance. However, prompt sensitivity and task interference remain challenges that future work must address to fully realize its potential as a general-purpose AI model.

## 3. GAPS AND CHALLENGES

VQA faces several critical challenges that impact its reliability, fairness, and generalization. These challenges span dataset limitations, biases, model robustness, and ethical concerns.[7][8]

### 3.1 Bias in VQA models:

- Language Bias: VQA models rely heavily on textual patterns rather than images, leading to poor generalization. They often predict answers based on statistical regularities in the dataset instead of truly understanding the visual content.
- Multimodal Bias: The frequent co-occurrence of specific words and visual elements causes models to memorize shortcuts rather than perform reasoning.
- Gender & Racial Bias: Datasets contain social biases, potentially leading to unethical or discriminatory outputs.

### 3.2 Dataset Limitations

- Imbalanced Datasets: Many datasets contain class imbalances, making it difficult for models to generalize.
- Image-Text Misalignment: Errors in dataset annotations lead to incorrect associations between images and text.
- Lack of Real-World Diversity : Many datasets fail to represent real-world complexities, leading to overfitting on training distributions.
- Out-of-Distribution (OOD) Weakness: VQA models struggle when tested on unseen image-question pairs, revealing limitations in reasoning capabilities.
- Cross-Lingual Limitations: Most datasets are English-centric, leading to performance drops in multilingual applications.

### 3.3 Model Evaluation Challenges

- Generalization & Robustness : High accuracy on training datasets does not indicate robustness, as models often fail on adversarial or real-world variations.
- Inconsistent Predictions: VQA models provide different answers for semantically similar questions, raising concerns about their reliability.
- Generative Answer Evaluation: Evaluating long-form or generative responses remains challenging due to answer style variations.

- Lack of Commonsense & External Knowledge Integration: VQA models struggle with reasoning that requires additional background knowledge.

### 3.4 Debiasing and Robustness Challenges

- Trade-off Between In-Distribution (ID) and OOD Performance: Debiasing techniques often improve OOD accuracy at the cost of ID performance, making models less effective on standard benchmarks.
- Limited Effectiveness of Debiasing Methods: While approaches like ensemble learning, data augmentation, and contrastive learning help, they do not fully solve the problem of statistical shortcuts.

### 3.5 Ethical and Computational Issues

- Fairness & Ethical Bias: Training on biased datasets can reinforce stereotypes, leading to harmful predictions.
- Privacy Risks: VQA applications in surveillance and biometric systems raise ethical concerns regarding user privacy.
- Computational Costs: Training robust VQA models requires extensive computational resources, raising concerns about sustainability.

## 4. CONCLUSION

Advanced VQA models employ a variety of approaches to address key challenges in cross-modal alignment, data efficiency, and reasoning, driving progress toward human-level visual understanding. ABC-CNN enhances visual attention by enabling the model to focus on relevant image regions based on the input question. KICNLE improves reasoning through knowledge augmentation and iterative refinement, ensuring coherent and explainable answer-explanation pairs. Masked Vision-Language modeling strengthens cross-modal representations by reconstructing masked signals, fostering a more robust multimodal understanding. Meanwhile, BLIP-2 prioritizes computational efficiency by leveraging frozen unimodal models with a lightweight querying transformer, achieving state-of-the-art results with significantly fewer trainable parameters. Lastly, OFA's unified architecture streamlines multimodal learning across tasks, seamlessly integrating vision, language, and cross-modal applications through a single sequence-to-sequence framework. Together, these models bring complementary strengths, tackling various aspects of the VQA challenge, from focused attention and knowledge-driven reasoning to efficiency and adaptability in multitask learning.

Beyond the models discussed, recent advancements in VQA include Large Vision-Language Models (VLMs) such as GPT-4, Gemini, and Claude, which enhance reasoning and compositional abilities. Additionally, Video Question Answering (VideoQA) leverages video frames and transcripts to improve temporal reasoning, while Neuro-Symbolic VQA integrates neural networks with symbolic logic for structured answer generation. These emerging methods further expand the capabilities of VQA systems.

**Future Directions:**

- Developing Fairer & More Diverse Datasets: Ensuring balanced and representative datasets to improve generalization and fairness, addressing biases that may affect real-world applications.

- Improving Evaluation Metrics: Creating better methods for assessing reasoning, consistency, and multimodal alignment, ensuring that VQA systems are judged by more comprehensive and interpretable benchmarks.
- Enhancing Energy-Efficient Models: Reducing computational overhead to make VQA more sustainable, enabling real-world deployment with lower energy consumption.
- Incorporating Time-Series Image Analysis: Moving beyond static images to enable models to predict subsequent frames or monitor environmental changes, benefiting applications like weather forecasting and autonomous navigation.
- Developing Lightweight, Efficient Models: Ensuring that VQA systems maintain high performance while being deployable on smaller devices, improving accessibility for real-world applications.
- Enhancing Commonsense Reasoning: Building on knowledge-augmented models like KICNLE, future VQA systems could improve contextual understanding and logical inference, benefiting fields such as medical diagnostics and autonomous systems.